\pdfoutput=1
\documentclass[11pt]{article}

\usepackage[final]{acl}

\usepackage{times}
\usepackage{latexsym}
\usepackage[english]{babel}

\usepackage[T2A, T1]{fontenc}
\usepackage[utf8]{inputenc}

\usepackage{microtype}

\usepackage{inconsolata}

\usepackage{graphicx}
\usepackage{subcaption}

\usepackage{multirow}
\usepackage{caption}
\usepackage{float}

\title{Matching and Linking Entries in Historical Swedish Encyclopedias}

\author{Simon Börjesson\footnotemark[1], Erik Ersmark\footnotemark[1], Pierre Nugues \\
        Lund University \\ Lund, Sweden \\
  \texttt{\{si7405bo-s, er5612er-s\}@student.lu.se, pierre.nugues@cs.lth.se}\\\\
\textit{Paper originally published in the Proceedings of the 9th Joint SIGHUM Workshop on} \\
\textit{Computational Linguistics for Cultural Heritage, Social Sciences, Humanities and Literature}\\ 
\textit{(LaTeCH-CLfL 2025)}}
\begin{document}
\maketitle

\begin{abstract}
The \textit{Nordisk familjebok} is a Swedish encyclopedia from the 19th and 20th centuries. It was written by a team of experts and aimed to be an intellectual reference, stressing precision and accuracy. This encyclopedia had four main editions remarkable by their size, ranging from 20 to 38 volumes. As a consequence, the \textit{Nordisk familjebok} had a considerable influence in universities, schools, the media, and society overall.
%with over ??XX?? copies sold. 
%Across the editions, the selection of entries and their content varied, reflecting the intellectual changes in Sweden.%
As new editions were released, the selection of entries and their content evolved, reflecting intellectual changes in Sweden.

% was present in all the public libraries of Sweden , 

In this paper, we used digitized versions from \textit{Project Runeberg}. We first resegmented the raw text into entries and matched pairs of entries between the first and second editions using semantic sentence embeddings. We then extracted the geographical entries from both editions using a transformer-based classifier and linked them to Wikidata. This enabled us to identify geographic trends and possible shifts between the first and second editions, written between 1876–1899 and 1904–1926, respectively.

Interpreting the results, we observe a small but significant shift in geographic focus away from Europe and towards North America, Africa, Asia, Australia, and northern Scandinavia from the first to the second edition, confirming the influence of the First World War and the rise of new powers. The code and data are available on GitHub at \url{https://github.com/sibbo/nordisk-familjebok}.
\end{abstract}

%\textbf{Keywords:} 
%knowledge extraction,
%text segmentation,
%entity linking,
%KB-BERT, 
%Wikidata

% vad hade man sökt på för att hitta den här? kb-bert kanske 

\renewcommand*{\thefootnote}{\fnsymbol{footnote}}
\footnotetext[1]{Equal contribution}
\renewcommand*{\thefootnote}{\arabic{footnote}}
\setcounter{footnote}{0}

\section{Introduction}
Encyclopedias are semi-structured, information-rich bodies of knowledge. In the field of knowledge extraction, their organization into articles with a headword makes them easier to process. 

Before the advent of the internet, major encyclopedias like the \textit{Encyclopædia Britannica}, \textit{Brockhaus Enzyklopädie}, and \textit{Nordisk familjebok} regularly released new printed editions for decades or even centuries. Largely written by academics and experts, %\citep{christensson2005i},
each edition reflects the knowledge base of the educated class in their respective region at that time. Through digitization efforts, many of these editions are available online.

The \textit{Nordisk familjebok} is widely recognized as the most comprehensive and influential Swedish encyclopedia \citep{aronsson2003preface,simonsen2016udgivelse}. The encyclopedia was published in four main editions between 1876 and 1993, with over 100 volumes and several hundred thousand articles. Starting in 2003, volunteers at \textit{Project Runeberg}\footnote{\url{https://runeberg.org/nf/}} scanned the paper volumes, applied an optical character recognition (OCR) to the images, and proofread a part of the entries.

Linking entries between editions to identify shared, added, and removed articles can indicate changes in the perception of information value or importance due to, e.g., world events or new technologies. One way of looking at this is the geographical spread of entries, i.e., if locations in some countries receive more or less attention over time. Linking entries to a graph database like Wikidata, which has coordinates listed for most entities tied to a location, can help highlight these trends.

% Finns inte mycket info om detta. Behöver låna boken som Axel verkar kollat på.

% \begin{enumerate}
    % \item \textit{Första utgåvan}, The first edition: 
    % \item \textit{Uggleupplagan}, The second edition:
    % \item \textit{Tredje upplagan}, The third edition:
    % \item \textit{Fjärde upplaga,}, The fourth edition:
% \end{enumerate}

The main contributions of our paper are:

\begin{enumerate}
    \item We scraped and segmented the first and second editions of the \textit{Nordisk familjebok} OCRed by \textit{Project Runeberg};
    \item We classified the segmented entries to identify the locations and cross-references;
    \item We matched pairs of entries between the two editions (first and second);
    \item We linked entries from both editions to unique Wikidata identifiers;
    \item We provide a brief interpretation of the changes in geographic focus from the first to the second edition.
\end{enumerate}

Our code is available on GitHub: \url{https://github.com/sibbo/nordisk-familjebok}.
%Our code will be made available on GitHub when this paper is accepted.

\section{Previous Work}
This work addresses three main problems: classifying entries, matching them across editions, and linking each entry to its counterpart in a knowledge graph like Wikidata. We outline relevant techniques and review previous work. Many of them use models trained on English. We also describe models specific to Swedish.

\subsection{Categorizing Entries}
In this work, we only considered entries describing a location. We extracted these entries using a supervised text categorization technique. \citet{Lewis2004} is an early example of such a technique with a large corpus, where the authors describe the annotation of one million newswires and baseline techniques to classify them.

CLD3\footnote{\url{https://github.com/google/cld3}} is a compact model created for language classification. It uses character $n$-grams as input to train a two-layer neural network model. Beyond language detection, CLD3 can be applied to other text classification tasks. 

The transformer architecture \citep{vaswani2017attention} with the BERT encoder component \citep{devlin-etal-2019-bert} reported state-of-the-art performances in the GLUE benchmark \citep{Wang2018} for classification tasks. Through language model pre-training, BERT achieves an impressive understanding of language, enabling it to grasp complex semantic and contextual nuances. It thus decreases the necessary amount of annotated samples to reach high classification scores.

%utilizes the encoder component of . The encoder is designed to capture contextual information between words. 

\subsection{Matching Entries}
Text matching refers to the quantification of the semantic similarity of a pair of documents, here encyclopedia entries. Applications of text matching include information retrieval and question answering. The TF-IDF document vectorization \citep{Sparck1972} is a baseline technique for representing documents, and the cosine similarity of two document vectors is a standard measure for evaluating their relatedness.

%Linking articles requires comparisons of sentence embeddings. In the context of \textit{Nordisk familjebok}, this task reaches an uncomfortable scale, as even the most basic methods would require comparing each encyclopedia entry with sentence embeddings from potentially over 100,000 articles. 

Dense vector representations of sentences or documents \citep{cordier-1965-factor} have proven to be better than sparse ones such as TF-IDF to encapsulate their semantics. \citet{reimers2019sentence} showed they could train transformer models from pairs of similar sentences and embed them in the form of dense vectors reflecting their semantic proximity. 

%While document embeddings and cosine similarity provide a ranking, in our setup, we match pairs, where the first pair is an entry in the first edition and the second one is an entry in the second edition. Linking pairs between editions 
In our setup, we want to match pairs of corresponding articles between editions, which requires comparing similarity scores of embeddings. In the context of the \textit{Nordisk familjebok}, the brute force method of comparing each article in one edition to all articles in the other quickly becomes unmanageable. With more than 100,000 articles per edition, this results in over $10^{10}$ comparisons.

Vector databases allow for much faster comparisons through efficient storage and indexing of vectors, employing algorithms like the hierarchical navigable small world algorithm and R-trees \citep{kukreja2023vector}. Vector databases can use SBERT models to vectorize the documents or more elaborate algorithms such as those of \citet{bge_embedding}, \citet{SFR-embedding-2}, or \citet{lee2024nvembedimprovedtechniquestraining}.

%?? Possibly see other recent models  here: \url{https://huggingface.co/spaces/mteb/leaderboard}

\subsection{Adapting Models to Swedish}
KB-BERT \citep{malmsten2020playing} is one of the Swedish BERT models developed at \textit{Kungliga biblioteket} (KB), the National Library of Sweden. It is trained on a corpus of Swedish texts created between 1940-2019, including newspapers, government publications, e-books, social media posts, Swedish Wikipedia, and more. Using a teacher-student model with KB-BERT as the student model, they also created a Swedish sentence transformer, KB-SBERT v2.0 \citep{rekathati2023swedish}.

\subsection{Linking Entries}

Wikidata is a free online knowledge graph containing over 115 million items at the time of this study\footnote{\url{https://www.wikidata.org/wiki/Special:Statistics}}. Each item has a unique QID and a number of property-value pairs that describe it. For example, Sweden's capital, Stockholm, has the QID \texttt{Q1754}, and its properties include \texttt{P625}, describing its coordinate location. 
%The \texttt{P625} property is particularly useful when linking entities to their location.

A few works have explored the task of linking named entities to Wikidata. \citet{shanaz2021wikidata} linked persons mentioned in newspapers, and \citet{nugues2022connecting} linked location entries from the French dictionary \textit{Petit Larousse illustré} to their corresponding coordinates in Wikidata. \citet{ahlin2024mapping} undertook a similar task to this study, linking location entries from the second edition of the \textit{Nordisk familjebok} to Wikidata. 

%Both \citet{nugues2022connecting} and \citet{ahlin2024mapping}, utilized the P625 property to link location entries to their respective coordinates in Wikidata.

\section{Preprocessing}
\label{sec:preprocessing}
\textit{Project Runeberg} is an online archive of old Scandinavian literature \citep{aronsson2023about}. This archive provides complete digital facsimiles and OCR texts of the first, second, and fourth editions of the \textit{Nordisk familjebok}, and parts of the third. Volunteers have carried out a manual proofreading on the vast majority of the OCR texts of the first edition, and parts of the second edition, as well as creating a currently incomplete index over the entry headwords on each page.

\subsection{Scraping}
We scraped the web pages of the first and second editions of the \textit{Nordisk familjebok} on the \textit{Project Runeberg} website, with the exception of the supplements.  
% Man kan skriva om Bok struktur i introduktionen. 
We parsed the HTML pages so that we could extract the index of entries on each page, extracted the raw OCR text, and finally removed or replaced most HTML tags and uncommon Unicode characters. 

\subsection{Segmenting}
% Utkast till inledning av Segmenting, kanske nämna att vi inte tog subentries.
The segmentation of the raw scraped text revealed a complex problem. While the entry headwords in the physical copies of the \textit{Nordisk familjebok} are always in bold characters, there is often no corresponding markup in the digitized text from \textit{Project Runeberg}, probably due to a rudimentary OCR conversion. This is especially true for the second edition, which at the time of this study had undergone less proofreading than the first edition. To deal with this, we devised a three-step approach: 
\begin{enumerate}
    \item \textbf{Bold matching}: If the paragraph begins with a bold tag, it is an entry.
    \item \textbf{Index matching}: Else, if the paragraph does not begin with a bold tag but starts with a headword present in the index, it is an entry.
    \item \textbf{Entry classification}: Otherwise, utilize a binary classifier model for entry classification.
\end{enumerate}
Following \citet{ahlin2024mapping}, who observed that excessively long texts negatively impacted the performance of their location classifier, we truncated entry texts to a maximum of 200 characters. 

%\citet{ahlin2024mapping} observed that excessively long texts negatively impacted the performance of their location classifier, speculating that it introduced irrelevant information. Therefore, we truncated the entry texts to a maximum of 200 characters. Additionally, we created a unique identifier (\textit{entryid}) for each entry. 

Some entries have numbered subentries under the same headword. This is notably the case with entries for noble lineages and royal houses, containing a list of people under the same family name, as for instance the \textit{Leijonhufvud}\footnote{\url{https://runeberg.org/nfai/0520.html}} and \textit{Natt och Dag}\footnote{\url{https://runeberg.org/nfbs/0318.html}} families.
%Since we were mainly concerned with geographical entries and we found few subentries about locations,\footnote{See e.g. \textit{Aachen} (\url{https://runeberg.org/nfba/0015.html}).} we did not consider subentries in this paper.
For sake of simplicity, we did not consider subentries in this paper.
%?? Do you have a example of one or two? ??

\subsubsection{Bold Matching}
We applied the rule that a paragraph is an entry if it begins with an HTML bold tag, \verb=<b>=. The headword is chosen as the text between the opening bold tag \verb=<b>= and the closing bold tag \verb=</b>=, removing any trailing punctuation.

\subsubsection{Index Matching}
The index contains the headwords of all entries on a page. They are manually added by proof-readers, which invariably gives rise to human errors. This, together with OCR errors, makes strict character comparisons of index words and entry texts impractical. 

We utilized the Levenshtein distance \citep{levenshtein1966binary} to match the index words to the raw text. We found that many of these index words were too long for absolute edit distance to fairly represent the similarity of these words. Therefore, we extended the Levenshtein distance metric to be relative to word length and, through manual testing, set a match threshold of 0.15. 

With these prerequisites, the method greedily attempts to match the longest index word to a substring of the same character length, starting at the beginning of the paragraph. In the event of a match, the index word is chosen as the entry headword.

% Kanske lite av detta ska stå i previous work eller inte förklaras alls.
%The \textit{Levenshtein edit distance} is a metric that measures the number of modifications required to transform one text into another. A modification is either a character insertion, removal, or replacement, all attributing one distance unit to the metric. 

\begin{figure*}[tb]
    \centering
    \includegraphics[width=\textwidth]{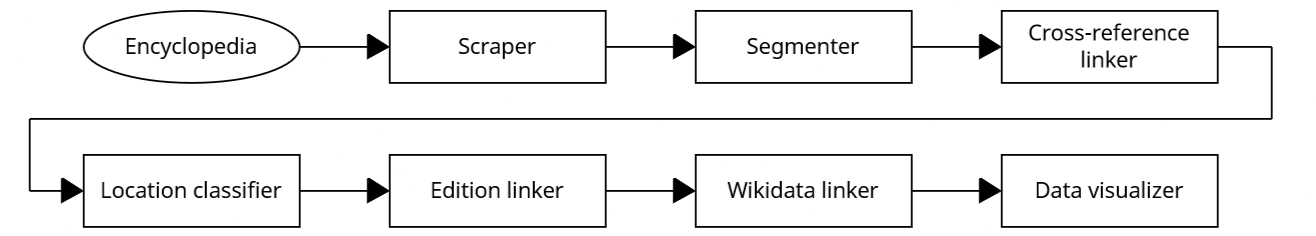}
    \caption{Overview of the pipeline.}
    \label{fig:pipeline}
\end{figure*}

\subsubsection{Entry Classification} 
% CLD3 och labben kan förklaras i previous work.

We created an entry classifier from a reimplementation of Google's CLD3 architecture. This provided us a foundation for a general classification model that is well-suited for exploiting small semantic details in the texts. 

Paragraphs in the scraped text that were indeed articles often contained distinctive features, such as punctuation and different types of parentheses. Therefore, we determined that a logistic head, instead of a two-layer network, would suffice for entry classification.

To create a training set, we leveraged the structure of the encyclopedias. Given that a paragraph beginning with a bold tag is almost certainly a valid entry, we used these paragraphs as ground truth for entries, removing bold tags in the process. Additionally, we used the fact that an encyclopedia is alphabetically ordered to find ground truth for non-entries. For example, in a volume, where all entries begin with the letter \textit{K}, a paragraph starting with any other capital letter is a non-entry. 

\subsection{Cross-references}
Many entries in the \textit{Nordisk familjebok} are cross-references, entries that refer to another entry and provide little to no information on their own, e.g.:
\begin{quote}
    \textit{\textbf{Nervtumör.} Se Nervsjukdomar.}\\
    ``\textit{\textbf{Nerve tumor.} See Neurological disorder.}''
\end{quote}

For the goals of the study, cross-references provided no value. 
%We also suspected that cross-references, if treated like other entries, would negatively impact subsequent steps. 
Therefore, we developed a rule to annotate an entry as a cross-reference if the text was shorter than 60 characters and contained the substring \texttt{\textvisiblespace Se} ``See''. We then extracted the word after \texttt{\textvisiblespace Se} and matched this word to an entry with that exact headword. Some cross-references are longer than 60 characters, but these entries usually provide some information on their own, so we left them as is.

\section{Method}
Figure~\ref{fig:pipeline} shows the processing pipeline consisting of scraping, segmenting, linking cross-references, location classification, edition linking, Wikidata linking, and data visualization.

We described the preprocessing modules, scraping, segmenting and linking cross-references in Section~\ref{sec:preprocessing}. In this section, we describe the rest of the architecture.

\subsection{Location Classifier}
To determine the location entries, we trained a binary classifier. We manually annotated 200 entries to create a training set of locations and non-locations. We used KB-BERT to tokenize the entry texts and encode them as in \citet{ahlin2024mapping}. We then fitted a logistic regression to the hidden states of the \verb=[CLS]= token. 
%?? Is that right?? %tokenization.

\subsection{Matching Pairs of Entries}
We matched the location entries of the first and second editions.
We created sentence embeddings of the entries with the KB-SBERT model and used a Qdrant vector database\footnote{\url{https://qdrant.tech/}} to store them. We then calculated the closest match using cosine similarity. For an entry from the first edition, we finally obtained a list of ranked candidates from the second. We used a greedy strategy and kept the first candidate.

Since always using the closest match leads to many false positives, especially for entries that only exist in one of the editions, we used a cosine similarity threshold value of 0.9 that maximized the F1 score on a manually annotated dataset of 200 entries.

This resulted in a list of matching pairs in the first and second editions of the \textit{Nordisk familjebok} as well as lists of removed and added entries.

%to find which articles remained, which were added, and which were removed from . 

\subsection{Wikidata Linking}
We linked entries marked as locations to Wikidata items and retrieved their geographical coordinates. This consisted of two steps: querying Wikidata and linking texts.

\subsubsection{Querying Wikidata} % Deras heter också detta :/
%For querying Wikidata, we utilized the \texttt{requests} package in Python with the Wikidata API.\footnote{API for querying Wikidata: \url{https://www.wikidata.org/w/api.php}} We queried Wikidata in Swedish for each entry headword and we chose the five top results.

%We utilized the Wikidata API,\footnote{\url{https://www.wikidata.org/w/api.php}} querying in Swedish with the entry headwords and choosing the five top results. For each Wikidata item, if there was a corresponding Swedish Wikipedia article, we retrieved the first 200 characters of it using the Wikipedia API.\footnote{\url{https://sv.wikipedia.org/w/api.php}} Otherwise, we used the Swedish Wikidata description. We designed our program to prefer Wikipedia texts, assuming that Wikidata descriptions were too concise for semantic comparisons. 

We queried the Wikidata API\footnote{
\url{https://www.wikidata.org/w/api.php}} with the entry headwords and chose the first five results. For each Wikidata item, we retrieved the first 200 characters of the corresponding Swedish Wikipedia article if available\footnote{Using the Wikipedia API: \url{https://sv.wikipedia.org/w/api.php}}. Otherwise, we used the Swedish Wikidata description. We designed our program to prefer Wikipedia texts, assuming that the more encyclopedic Wikipedia text would better match the entry texts. 

%We suspected that the Wikipedia text would result in more accurate semantic comparisons, since the Wikidata descriptions are shorter and more general. Therefore, we designed or program to prefer Wikipedia texts. 

\subsubsection{Linking Texts}
We encoded the segmented entry text and the retrieved texts of each Wikidata item with the KB-SBERT model, and we compared the encyclopedia entry to each item to find the highest cosine similarity score. Due to the limited search space of five items, we extended the method with a matching threshold, chosen through evaluation on two test sets consisting of 25 random locations from each edition, respectively. We achieved the best F1 scores with a threshold of 0.6.

%We extended the method with a matching threshold because a search space of five items is quite limited, potentially leading to false positives. We chose a threshold by evaluating the method on two test sets, one for each edition, each consisting of 25 randomly selected locations from the segmentation. We achieved the best F1-scores with a threshold of 0.6. 

%Due to the limited search space of five items, we extended the method with a matching threshold of 0.6, chosen through evaluation on two test sets consisting of 25 random locations from each edition respectively.

%We compared the vectors using the cosine similarity method, finding the closest match, and disregarding it if the similarity score was less than a threshold of 0.6. 
%We deemed it necessary to have a threshold since a search space of 5 items is quite limited, and could otherwise significantly increase the number of false positives. We chose this score by evaluating the method on 2 test sets, each consisting of 25 randomly selected locations from the segmenations of the first and second editions. We achieved the best F1-scores with a threshold of 0.6. 

Lastly, we retrieved the QID and the geographical coordinates using the coordinate location property (\verb=P625=) of the best match that passed the threshold. % Setting the entity attributes to null if no result for either. 

\begin{table}[b]
  \centering
  \begin{tabular}{lrlll}
    \hline
    \textbf{Ed.} & 
    \textbf{Entries}  & \textbf{Bold} & \textbf{Index} & \textbf{Classifier} \\
    \hline
    $1^{st}$ & 84,534   & 97.7\%        & 2.14\%         & 0.17\%             \\
    $2^{nd}$ & 150,340  & 76.0\%        & 11.5\%         & 12.5\%             \\
    \hline
  \end{tabular}
  \caption{\label{tab:segmenter}
    The total number of entries segmented for both editions, and the proportion of entries segmented using each of the three strategies.
  }
\end{table}

\begin{table}[b]
  \centering
  \begin{tabular}{lrrr}
    \hline
    \textbf{Ed.} & 
    \textbf{Entries}    & \textbf{Locations}& \textbf{Proportion}   \\
    \hline
    $1^{st}$ & 84,534   & 18,932            & 22.4\%                \\
    $2^{nd}$ & 150,340  & 32,378            & 21.6\%                \\
    \hline
  \end{tabular}
  \caption{\label{tab:location-classifier}
    The total number of entries segmented for both editions, the number of entries classified as locations, and the corresponding proportions.
  }
\end{table}

\begin{table*}[!t]
\centering
\begin{tabular}{lllllll}
\hline
\multirow{2}{*}{\textbf{Method}}     & \multicolumn{3}{c}{\textbf{First edition}} & \multicolumn{3}{c}{\textbf{Second edition}}                \\
& Precision   & Recall  & F1-score  & Precision & Recall & F1-score \\
\hline
\multicolumn{1}{l|}{Segmenter, weighted mean$^2$}
& $\approx$1.0& 1.0     & \multicolumn{1}{l|}{1.0}         & 0.99       & 0.94      & 0.96     \\
\multicolumn{1}{l|}{\textit{\ \ Bold matching}$^2$}
& 1.0         & 1.0     & \multicolumn{1}{l|}{1.0}          & 1.0       & 1.0       & 1.0      \\
\multicolumn{1}{l|}{\textit{\ \ Index matching}$^2$}
& 0.96        & -       & \multicolumn{1}{l|}{-}            & 0.94      & -         & -        \\
\multicolumn{1}{l|}{\textit{\ \ Entry classifier$^4$}}
& 0.95        & 0.95    & \multicolumn{1}{l|}{0.95}         & *         & *         & *        \\
\multicolumn{1}{l|}{Cross-references$^3$}
& 1.0         & 0.85    & \multicolumn{1}{l|}{0.92}         & 1.0       & 0.75      & 0.86     \\
\multicolumn{1}{l|}{Location classifier$^1$}
& 0.84        & 0.96    & \multicolumn{1}{l|}{0.90}         & 0.92      & 0.92      & 0.92     \\
\multicolumn{1}{l|}{Entry matching$^4$}
& 0.85        & 0.83    & \multicolumn{1}{l|}{0.83}         & *         & *         & *        \\
\multicolumn{1}{l|}{\textit{\ \ Baseline: headword match}$^3$}
& 0.74         & 0.81     & \multicolumn{1}{l|}{0.76}          & *       & *       & *      \\
\multicolumn{1}{l|}{Wikidata linking}
&             &         & \multicolumn{1}{l|}{}             &           &           &          \\
\multicolumn{1}{l|}{\textit{\ \ QID match$^1$}}
& 0.40        & 0.52    & \multicolumn{1}{l|}{0.45}         & 0.48      & 0.16      & 0.24     \\
\multicolumn{1}{l|}{\textit{\ \ Within 25 km$^1$}}
& 0.76        & 0.64    & \multicolumn{1}{l|}{0.69}         & 0.84      & 0.40      & 0.54     \\
\hline
\end{tabular}
\caption*{
\textsuperscript{1} 25 entries used, \textsuperscript{2} 50 entries used, \textsuperscript{3} 100 entries used, \textsuperscript{4} Used respective training/test data, '\textbf{-}' : The metric was not applicable, '\textbf{*}' : The values are the same for both editions.}

\caption{\label{tab:performance}
Performance metrics of the pipeline for both editions}
\end{table*}

\begin{figure*}[!t]
    \begin{center}
    \begin{minipage}[b]{0.491\textwidth}
        
        \includegraphics[width=\textwidth]{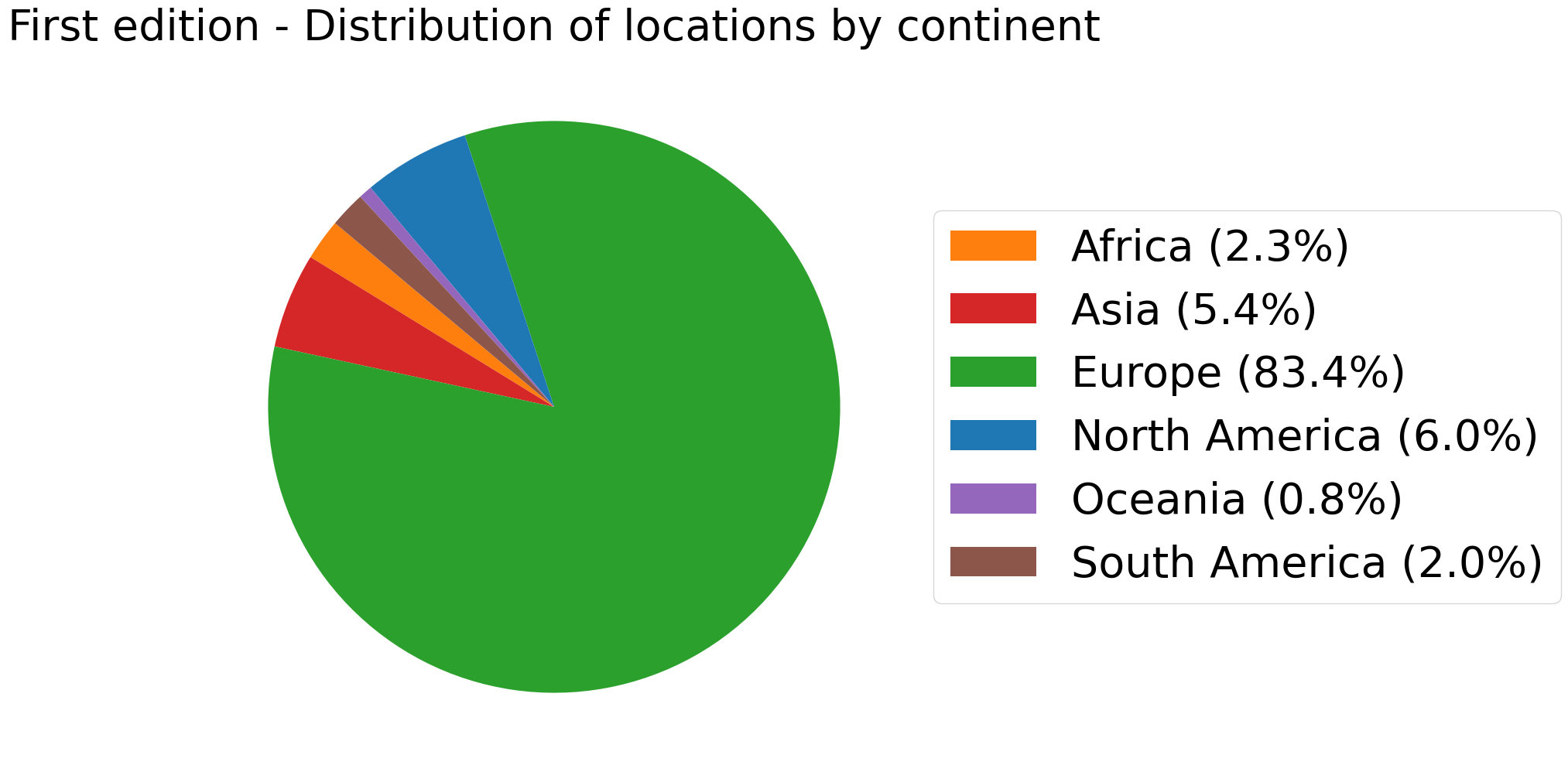}
        \subcaption{Distribution of locations by continent in the first edition.}
        \label{fig:e1-circle}
    \end{minipage}%
    \begin{minipage}[b]{0.509\textwidth}
        
        \includegraphics[width=\textwidth]{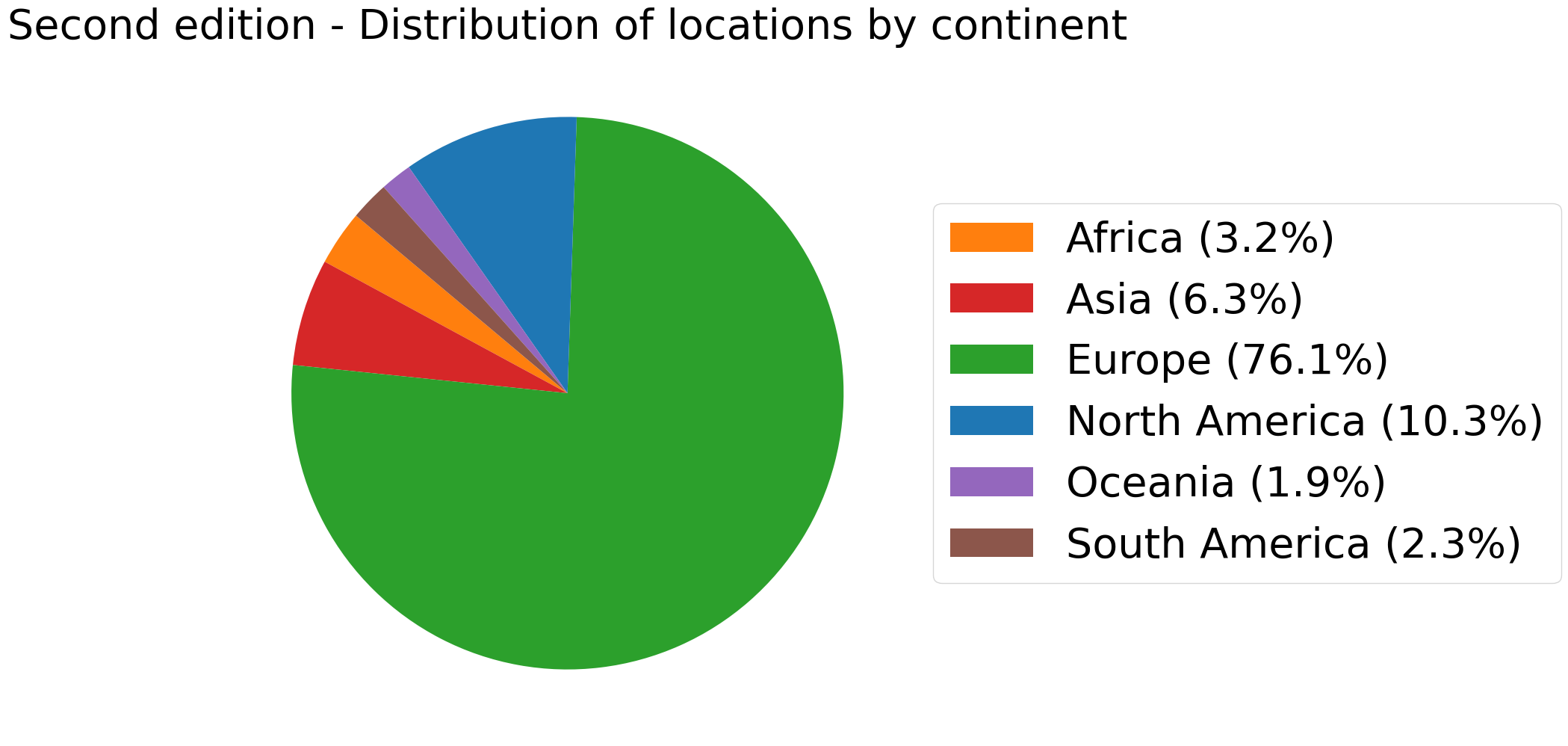}
        \subcaption{Distribution of locations by continent in the second edition.}
        \label{fig:e2-circle}
    \end{minipage}
    \end{center}
    \vspace{1em}
    \begin{center}
    \begin{minipage}[b]{0.95\textwidth}
        \begin{center}
        \includegraphics[width=\textwidth]{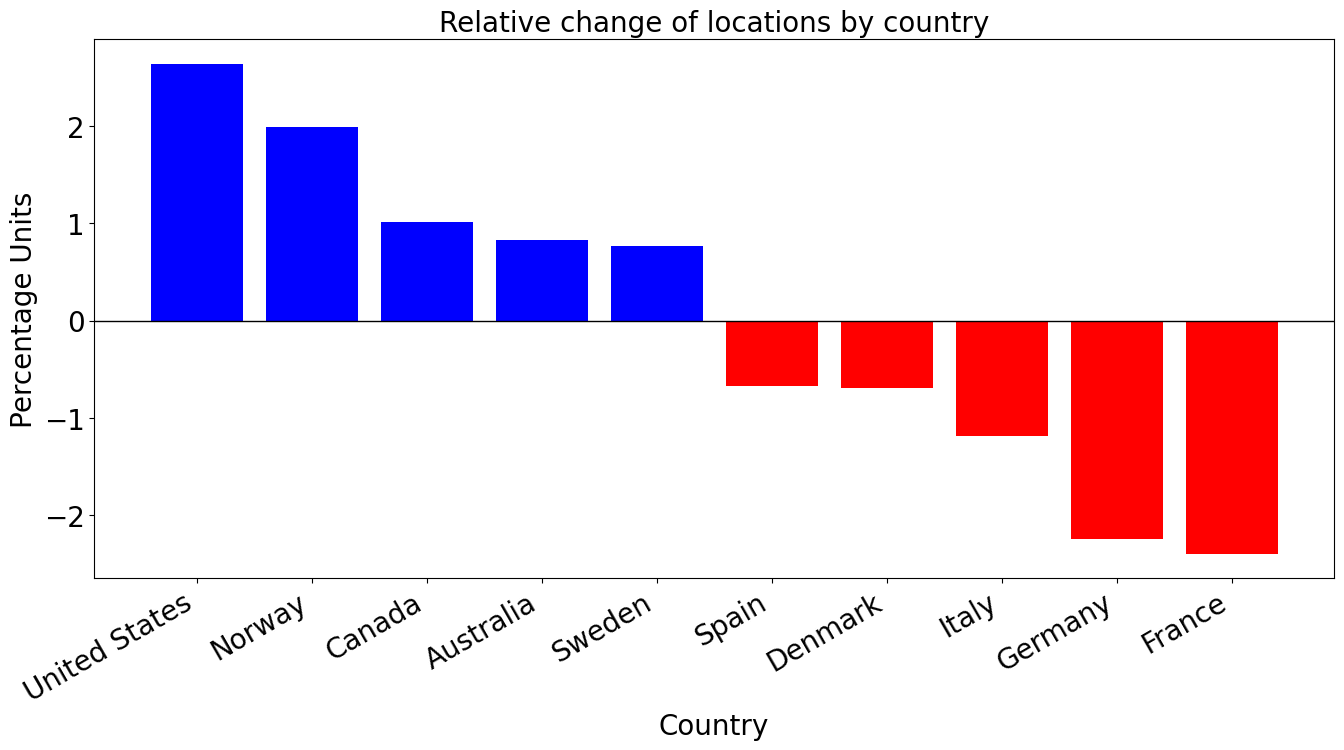}
        \end{center}
        \subcaption{The top five countries with the largest percentage unit increase (blue), top five countries with the largest percentage unit decrease (red), in location counts from the first edition to the second edition.}
        \label{fig:bar-plot}
    \end{minipage}
    \end{center}
    \caption{Location-related statistics from both editions.}
    \label{fig:loc-stats}
\end{figure*}

\begin{figure*}
    \centering
    \includegraphics[width=0.95\textwidth]{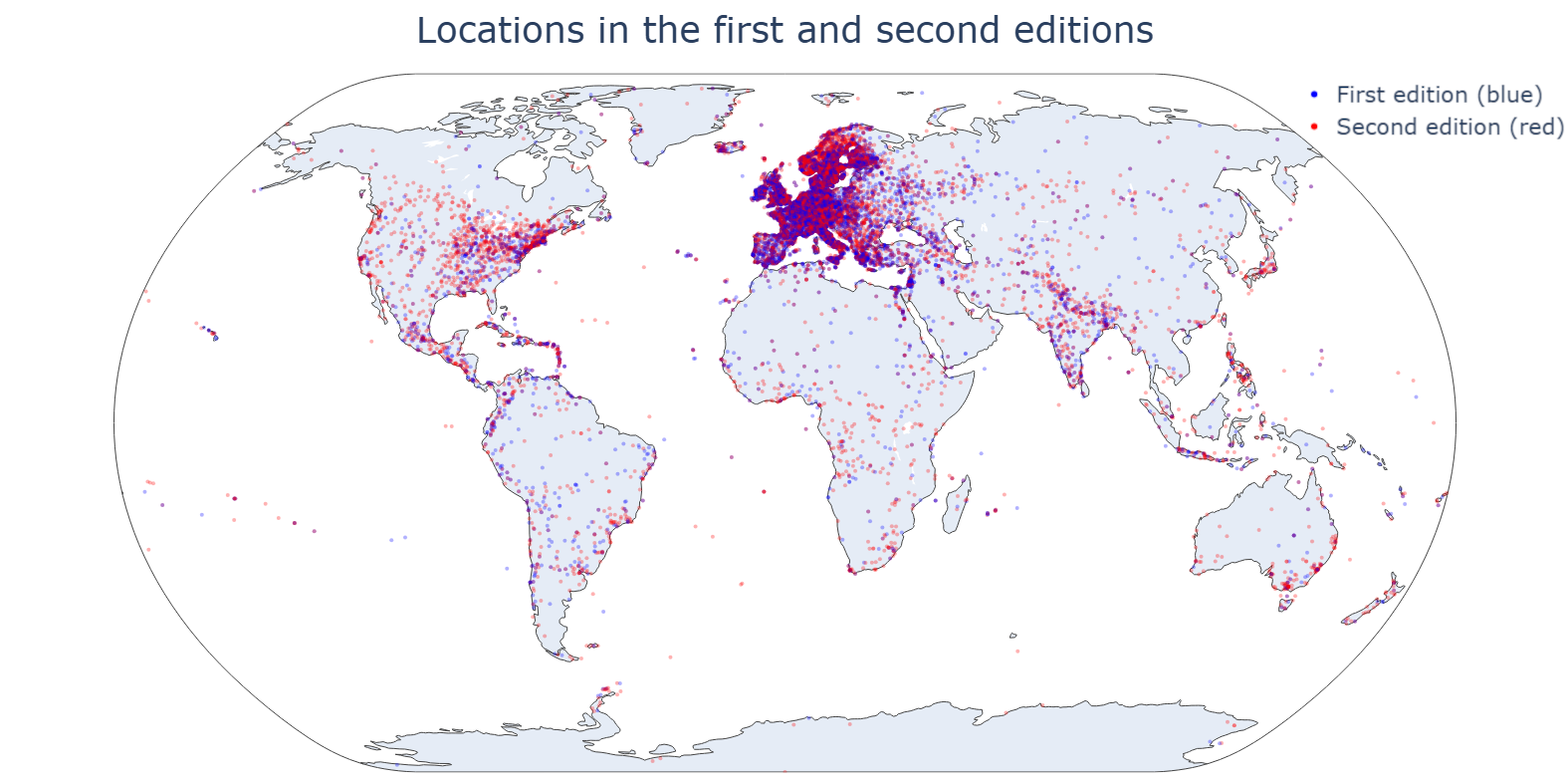}
    \caption{Geographic distribution of locations in both editions.}
    \label{fig:world-map}
\end{figure*}

\section{Results and Evaluation}
Table~\ref{tab:performance} shows the precision and recall scores of all parts of the pipeline where applicable. Most precision and recall scores were acquired by evaluating validation sets of 25, 50 or 100 random entries either in the encyclopedias or in the JSON files. These validation sets should give a general idea of the performance of each part. Nonetheless, their size is relatively small and larger sets would certainly improve their reliability and statistical significance.

\subsection{Segmenter}

In Table~\ref{tab:segmenter}, we can see that the second edition has roughly double the number of entries compared to the first one. The number of matches we obtained with the index and classifier strategies is very low in the first edition, since it has been proofread almost completely. 

\citet{christensson2005i} estimates the number of entries in the first edition to 103,000. The disparity between this and our 84,534 entries is likely due to not segmenting supplemental volumes. 

\citet{ahlin2024mapping} report the extraction of 130,383 entries when segmenting the second edition, while \citet{simonsen2016udgivelse} estimates over 182,000 headwords. Both included supplemental volumes, which we chose to exclude, but like us, they also omitted subentries. We believe the difference from the former is due to using index matching and a binary classifier for entries without bold tags, and the discrepancy from the latter again is mainly due to not segmenting the supplemental volumes.

In combination with the recall and precision scores for segmenting in Table~\ref{tab:performance}, we can be relatively certain that these numbers are good estimates for the total number of entries in the encyclopedias, excluding subentries and supplemental volumes.

\subsection{Cross-references}
%?? I am not sure I understsand the next paragraph. Can you try to reformulate it. Do not use negations??

%Since our goal was not to find all cross-references, but only those under a certain length that did not contribute enough information on their own, the metrics in Table \ref{tab:performance} show the performance of actually linking them to their referenced entry. 

Table~\ref{tab:performance} shows the performance of linking cross-references to their referenced entry. The method was quite simple, and gave rise to some errors, most notably linking the cross-reference to an incorrect entry with the same headword. For example, in the second edition, \textit{Bajesid} is listed as an alternate spelling of a lineage of sultans in the Ottoman Empire:
\begin{quote}
    \textit{\textbf{Bajesid,} turkiska sultaner. Se Bajasid.}\\
    ``\textit{\textbf{Bajesid,} Turkish sultans. See Bajasid.}''
\end{quote}

However, when trying to find the referenced entry \textit{Bajasid}, another cross-reference for the city of the same name is matched:
\begin{quote}
    \textit{\textbf{Bajasid,} stad. Se Bajaset.}\\
    ``\textit{\textbf{Bajasid,} city. See Bajaset.}''
\end{quote}

This is because the first entry with an exact headword match is chosen. For the purpose of removing redundant entries, we believe the performance of our method is satisfactory, but it could probably be improved by using a named entity recognizer.

\subsection{Location Classifier}
In Table~\ref{tab:location-classifier}, the ratio of locations in both editions is very similar, and the ratio in the second edition is almost identical to that of \citet{ahlin2024mapping} (21.7\%), which is expected since the same method was used. 

Table~\ref{tab:performance} shows the F1 scores of the location classifier for both editions. We can see that they match or surpass 0.9, which is notable considering the KB-BERT model was not fine-tuned for this task.

\subsection{Matching Entries}
The performance metrics presented in Table~\ref{tab:performance} demonstrate that our matching approach performs better than the baseline model (headword match) across all metrics, albeit not significantly. We had anticipated a more pronounced performance improvement from the more advanced KB-SBERT model compared to the simple baseline model. 

By examining matched sentences, it becomes apparent why certain errors occur. For instance, our method erroneously matched the following two entries, \textit{Åker} and \textit{Åsenhöga}:
\begin{quote}
    \textit{\textbf{Åker.} 1. Socken i Jönköpings län, Östbo härad. Areal 15,842 har. 1,798 innev. (1892). Å. bildar med...}\\
    ``\textit{\textbf{Åker.} 1. Parish in Jönköping county, Östbo hundred. Acreage 15,842 ha. 1,798 res. (1892). Å. forms with...}''
\end{quote} 
and
\begin{quote}   
    \textit{\textbf{Åsenhöga,} socken i Jönköpings län, Mo härad. 12,960 har. 1,257 inv. (1921). Å. bildar med...}\\
    ``\textit{\textbf{Åsenhöga,}  estate in Jönköping county, Mo hundred. 12,960 ha. 1,257 res. (1921). Å. forms with...}''
\end{quote} 

These entries exhibit strikingly similar semantic structures, with comparable word sets, order, and article topic. Scenarios like these are understandably difficult, and frequently occur in the corpus.

\subsection{Wikidata Linking}
When linking an entry to Wikidata, the best cosine similarity match was often not with the correct entity, but with a place or object not very far away, usually within only a few kilometers. A common error was matching a \textit{socken}, an old Swedish term for a church parish, to a nearby city, municipality, or building with the same or a very similar name. For example, 

\begin{quote}
\textit{\textbf{Öved}, socken i Malmöhus län...} \\
``\textit{\textbf{Öved}, parish in Malmöhus county...}''
\end{quote}
was linked to \textit{Övedsklosters slott}, a castle within the borders of the parish. 

%It is difficult to comprehend why this match in particular yielded the highest cosine similarity score, but linking errors like these make little to no difference on a global scale. Therefore, we decided to create a metric measuring if the matched Wikidata entity was within 25 kilometers of the correct coordinates. With this metric, the performance significantly improved for both editions, especially the second.

%While the results in Table \ref{tab:performance} are still quite poor, we assume that the spread of locations roughly follows the trends present in the respective edition, i.e., even though only around half of all locations present in the encyclopedias were linked to within 25 km of their correct coordinates, the distribution of locations should still be more or less the same.

It is difficult to understand why this match yielded the highest cosine similarity score, but such linking errors make little difference on a global scale. Therefore, we created a metric to check if the matched Wikidata entity was within 25 kilometers of the correct coordinates. Although this metric significantly improved performance for both editions, especially the second, the results in Table~\ref{tab:performance} remain quite poor. Even though only about half of all locations in the encyclopedias were linked within 25 km of their correct coordinates, it seems reasonable to assume that the overall distribution of locations remains roughly the same.

In Figure~\ref{fig:loc-stats}, we see a slight shift in focus away from large European countries like France, Germany, and Italy, towards primarily North America, Australia, Norway, and Sweden. We provide a brief interpretation of this in Section~\ref{sec:geographic-focus}.

%Another source of error stems from the limitation of search space we set for our method as we had to reduce computation time. This limitation had a significant impact for entries which are close to unknown. 

Another source of error stems from the limited search space we set to reduce computation time, which occasionally caused the program to miss the correct Wikidata item.

%Though an uncommon entry may exist in Wikidata, it is far from guaranteed to be included in the search results. When manually annotating validation sets, it could take us several minutes to find certain entries, e.g., the former \textit{Province of Brandenburg} in Prussia (\texttt{Q700264}) and the Russian \textit{Migulinskaya} (\texttt{Q7224942}). It is only when we used the Russian word in Cyrillic characters, \foreignlanguage{russian}{Мигулинская}, that we were able to find the latter. 

%The Wikidata search tool is limited; certain character modifications in a query can make or break a search. One example is \textit{Qvenneberga} parish, changed to \textit{Kvenneberga} in the second edition following Swedish spelling changes around the turn of the 19th century, among them changing the letter \textit{q} to \textit{k} in most words \cite{pettersson2005svenska}. The first headword yields no search results in Wikidata, while the second headword has several results.

The search functionality in Wikidata can be unreliable, especially for uncommon entries. For instance, finding the Russian location \textit{Migulinskaya} required using Cyrillic characters. Additionally, Sweden introduced a spelling reform around the turn of the 19th century. Among the changes was replacing the letter \textit{q} with \textit{k} in most words \citep{pettersson2005svenska}. For example, \textit{Qvenneberga} in the first edition became \textit{Kvenneberga} in the second one. Such small spelling changes can be crucial: The first term yielded no search results, while the second one resulted in a few hits. Altogether, these quirks can lead to search results missing valid entries, complicating the process of finding specific items.

\section{Discussion}
\subsection{Applications of Entry Matching}
The potential applications of matching entries across the editions of the \textit{Nordisk familjebok} are significant, especially in the context of digitization and preserving the relevance of this cultural artifact.
%for many Swedes.

One potential application is the development of a search system based entirely on the editions of the \textit{Nordisk familjebok}. This concept is currently being explored at the Centre for Digital Humanities at Gothenburg University.\footnote{\url{https://nordiskfamiljebok.dh.gu.se/}} Such a system could greatly benefit from the inter-edition links developed in this work, enabling comprehensive search results across all editions from a single query. 

Another application of our pipeline that could improve the accessibility of historical encyclopedias in the digital age is to extend Wikipedia pages with links to corresponding entries in digital facsimiles of encyclopedias.
%However, we recognize the need for further refinement of our method before it can be integrated into a search system for use by the general public.

\subsection{Geographic Focus}
\label{sec:geographic-focus}

%Outline results, time era, introduce reasons.
%As the world has a undergone rapid globalization since the industrialization, we expected a more even spread across continents in the second edition. As depicted in figures \ref{fig:e1-circle} and \ref{fig:e2-circle}, the geographic focus is indeed less centered on Europe in the second edition. 
Given the rapid globalization since the first edition,
we expected a more even geographic distribution in the second edition due to its later publication date. Figures~\ref{fig:e1-circle} and \ref{fig:e2-circle} confirm this hypothesis. The historical events that unfolded during the publication time frame of the editions could illuminate the reasons behind the observed changes.

%First world war 
% Källor. Jag har antingen en källa för varje land, eller en för alla. vad ska vi välja?
% https://www.awm.gov.au/articles/atwar/first-world-war
%The First World War involved many countries worldwide. Canada and Australia, both dominions of the British Empire, participated from the beginning of the war in 1914 when Britain declared war against Germany. The United States involvement in the war transpired between 1917 and 1918. The war also impacted Africa, with several regions under European occupation at the time. 
The First World War involved many countries worldwide, including Canada, Australia, the United States, Japan, and various European colonies in Africa. The involvement of these regions in the war may have influenced Swedish societal discourse, consequently affecting the content of the second edition \citep{ww1All}.

%Industrialization 
Figure~\ref{fig:bar-plot} shows an increase in the number of locations situated in Norway and northern Sweden. 
%From the late 19th century to the mid-20th century, Norway underwent significant industrialization, particularly in capitalizing on hydroelectric power \citep{norwayIndustrialization}. The northern parts of Sweden experienced a similar development between 1850 and 1950, primarily in the timber industry \citep{swedenIndustrialization}. Both of these processes led to population increases in these regions during that time, which may explain the new additions in the second edition. 
From the late 19th century to the mid-20th century, Norway and northern Sweden underwent significant industrialization in hydroelectric \citep{norwayIndustrialization} and timber production \citep{swedenIndustrialization}, respectively. Consequently, the population of these regions increased, which may explain these additions in the second edition.

Furthermore, Figures~\ref{fig:bar-plot} and \ref{fig:world-map} depict a relative decrease of location mentions for several European countries in the second edition. However, since the second edition contains more locations overall, it does not imply that the absolute number of location mentions has decreased for these countries.

\section{Conclusion}
In this paper, we compared two editions of a historical Swedish encyclopedia. We described the corpus collection, the segmentation of the raw text input into entries, the categorization of entries, and how we matched pairs of entries between the two editions. We finally reported how we linked geographical entries from both editions to Wikidata. 

In the classification and matching tasks, we used transformer models with parameters pre-trained on modern Swedish. A possible improvement is to fine-tune the models on older Swedish texts. We could also explore alternative algorithms for matching entries, such as the Hungarian algorithm \cite{kuhn1955hungarian}.

This work enabled us to identify shifts between the two editions and a few geographic trends. Most notably, the second edition reflects the evolution of the geographic awareness toward a more diverse global outlook. Beyond the historical events mentioned in Section~\ref{sec:geographic-focus}, there may be countless societal, cultural, political, and economic factors contributing to these changes. We hope our work will invite further investigation to provide a better understanding of the context surrounding them.

%\section{Future Work}
%Some potential future work based on this paper include:
%\begin{itemize}
%    \item Using a transformer model for binary classification of potential encyclopedia entries when the raw data is inconsistent;
%    \item Fine-tuning KB-BERT for older Swedish;
%    \item Normalizing older Swedish to fit models trained on modern text as in \citet{pettersson2016spelling};
 %   \item Fine-tuning KB-BERT for sentence pair classification when linking entries between editions and when linking to Wikidata;
%    \item Explore alternative algorithms for linking entries between editions, such as the Hungarian algorithm \cite{kuhn1955hungarian}; 
%    \item Querying Wikidata using Google or with some type of neural search.
%\end{itemize}

\section*{Limitations}
Our evaluation of headword detection and entry matching is limited and a comprehensive study would include more data. Our validation sets should give a general idea of the performance of each part. Nonetheless, their size is relatively small and larger sets would certainly improve their reliability and statistical significance.

Large language models that we used in this research may generate classification errors or show bias. This bias may come from the corpus used for training the models, mostly contemporary Swedish, while we applied them to the \textit{Nordisk familjebok} that uses a slightly different language.

\section*{Ethics Statement}
We identified a few potential risks:
\begin{enumerate}
\item The \textit{Nordisk familjebok} belongs to book history. It sometimes includes old-fashioned viewpoints and its information is dated.
\item This encyclopedia was written in a different historical context. A few entries may include content that can now be considered offensive. Potential users of our work or of applications based on it must be aware of this context.
\end{enumerate}

\section*{Acknowledgments}
We would like to thank the anonymous reviewers for their suggestions and comments.

This work was partially supported by \textit{Vetenskapsrådet}, the Swedish Research Council, registration number 2021-04533.

\bibliography{references}

\end{document}